\documentclass{article} 
\usepackage{iclr2025_conference,times}

\usepackage{amsmath,amsfonts,bm}

\def\eqref#1{equation~\ref{#1}}

\def\1{\bm{1}}

\def\rmE{{\mathbf{E}}}

\def\rmP{{\mathbf{P}}}

\def\evx{{x}}

\DeclareMathAlphabet{\mathsfit}{\encodingdefault}{\sfdefault}{m}{sl}
\SetMathAlphabet{\mathsfit}{bold}{\encodingdefault}{\sfdefault}{bx}{n}

\def\sA{{\mathbb{A}}}

\def\sT{{\mathbb{T}}}

\usepackage{bm}
\usepackage{hyperref}
\usepackage{url}
\usepackage{multirow}
\usepackage{graphicx}
\usepackage{algorithmicx}
\usepackage{algorithm}
\usepackage{algpseudocode}
\usepackage{wrapfig}
\usepackage{multirow}
\usepackage{amsmath}
\usepackage{amssymb}
\usepackage{booktabs}
\usepackage{array} 
\usepackage{adjustbox}
\usepackage{bbding}
\usepackage{ulem}
\usepackage{makecell}
\usepackage{enumitem}
\usepackage{colortbl}
\usepackage{soul, color, xcolor}
\soulregister{\cite}7 
\soulregister{\citep}7 
\soulregister{\citet}7 
\soulregister{\ref}7
\soulregister{\pageref}7 

\title{Retrieval Augmented Diffusion Model for Structure-informed Antibody Design and Optimization}

\author{%
  Zichen Wang$^{1}$\thanks{Equal Contribution} \ \ \ \ \ \ \ \ Yaokun Ji$^{1*}$ \ \ \ \ \ \ \ \ Jianing Tian$^{2}$\ \ \ \ \ \ \ \ Shuangjia Zheng$^{1}$\thanks{Corresponding author}
    \\
    $^1$ Global Institute of Future technology, Shanghai Jiao Tong University; \\
    $^2$ School of Software \& Microelectronics, Peking University \\
    \\
    \href{mailto:zichen\_w@sjtu.edu.cn}{\texttt{zichen\_w@sjtu.edu.cn}},  \href{mailto:shuangjia.zheng@sjtu.edu.cn}{\texttt{shuangjia.zheng@sjtu.edu.cn}}
}

\iclrfinalcopy 
\begin{document}

\maketitle

\begin{abstract}
Antibodies are essential proteins responsible for immune responses in organisms, capable of specifically recognizing antigen molecules of pathogens. Recent advances in generative models have significantly enhanced rational antibody design. 
However, existing methods mainly create antibodies from scratch without template constraints, leading to model optimization challenges and unnatural sequences.
To address these issues, we propose a retrieval-augmented diffusion framework, termed RADAb, for efficient antibody design. Our method leverages a set of structural homologous motifs that align with query structural constraints to guide the generative model in inversely optimizing antibodies according to desired design criteria. 
Specifically, we introduce a structure-informed retrieval mechanism that integrates these exemplar motifs with the input backbone through a novel dual-branch denoising module, utilizing both structural and evolutionary information. Additionally, we develop a conditional diffusion model that iteratively refines the optimization process by incorporating both global context and local evolutionary conditions.
Our approach is agnostic to the choice of generative models. Empirical experiments demonstrate that our method achieves state-of-the-art performance in multiple antibody inverse folding and optimization tasks, offering a new perspective on biomolecular generative models.
\end{abstract}

\section{Introduction}

Antibodies, essential Y-shaped proteins in the immune system, are pivotal for recognizing and neutralizing specific pathogens known as antigens. This specificity primarily arises from the Complementarity Determining Regions (CDRs), which are crucial for binding affinity to antigens \citep{jones1986replacing,ewert2004stability,xu2000diversity,akbar2021compact}. The design of effective CDRs is therefore central to developing potent therapeutic antibodies, a dominant class of protein therapeutics. 
However, the development of these antibodies typically relies on labor-intensive experimental methods such as animal immunization or screening extensive antibody libraries, often failing to produce antibodies that target therapeutically relevant epitopes effectively. Thus, the ability to generate new antibodies with pre-defined biochemical properties in silico carries the promise of speeding up the drug design process.

Computational efforts in antibody design have traditionally involved grafting residues onto existing structures \citep{sormanni2015rational}, sampling alternative native CDR loops to enhance affinities\citep{aguilar2022fragment}, and using tools like Rosetta for sequence design improvements in interacting regions \citep{adolf2018rosettaantibodydesign}. Many recent studies have focused on applying deep generative models to design antibodies \citep{luo2022antigen,martinkus2024abdiffuser,zhuabx}.
They take advantage of geometric learning and generative models to capture the higher-order interactions among residues directly from the data. These innovations provide more efficient methods to search sequence and structure spaces.

Albeit powerful, current generative models struggle to design antibodies that adhere to structural constraints and exhibit desired biological properties. This challenge primarily arises from a lack of diversity in the available training data. Predominantly, research efforts have relied on the SAbDab database \citep{dunbar2014sabdab}, which comprises fewer than ten thousand antigen-antibody complex structures. The limited scope of this dataset restricts the models' ability to capture comprehensive high-order interaction information between antigen-antibody residues, thereby increasing the risk of overfitting. Moreover, most existing methodologies attempt to design antibody sequences \textit{de novo}, without the benefit of template-based guidance. This approach inherently demands a greater volume of data and extensive training or fine-tuning on specific datasets to achieve efficacy in practical applications. 

\begin{wrapfigure}[13]{r}{0.5\textwidth}
        \centering
        \vspace{-4mm}
        \includegraphics[width=0.5\textwidth]{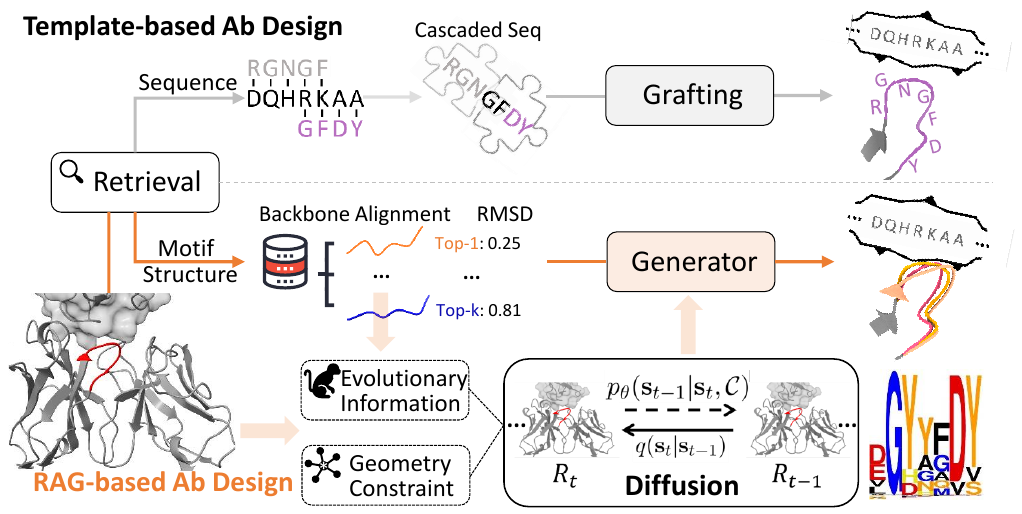}
        \vspace{-3mm}
        \caption{Illustration of the retrieval-augmented framework.}
        \label{illustration}
\end{wrapfigure}

In this work, we draw inspiration from template-based and fragment-based antibody design to develop a model that fully utilizes protein structural database, effective motif retrieval, and semi-parametric generative neural networks. Our goals are to: (a) leverage template-aware local and global protein geometric information to enhance model generative capability, (b) integrate motif evolutionary signals to prevent overfitting, and (c) necessitate minimal training or fine-tuning for effective use in real-world applications. 

To this end, we introduce the \textbf{R}etrieval-\textbf{A}ugmented \textbf{D}iffusion \textbf{A}nti\textbf{b}ody design model (RADAb), a novel semi-parametric antibody design framework. To fully exploit the protein structure space, we first compiled a database of CDR-like fragments from the non-redundant Protein Data Bank (PDB) \citep{berman2000protein}. These CDR-like fragments are linear functional motifs structurally compatible with an antibody CDR loop, found in any protein geometry within the PDB. For a given antibody to be improved, we perform a structural retrieval to obtain motifs with structures similar to the desired CDR framework. As protein sequences capable of folding into similar structures often share homology and consensus, we hypothesize that these retrieved motifs, enriched with evolutionary information, can enhance the model's generalization.

Unlike traditional rational design methods that optimize by grafting a single CDR fragment, we propose to use a set of structural homologous CDR-like motifs together with the desired backbone for iterative sequence optimization  (Figure \ref{illustration}). Our major contributions follow: (1) We propose a \textbf{first-of-its-kind} retrieval-augmented generative framework for rational antibody design. It uses a set of functional CDR-like fragments that satisfy the desired backbone structures and properties to guide generation toward satisfying all the required properties. (2) \textbf{A novel retrieval mechanism} is introduced for integrating these exemplar motifs with the input backbone through a novel dual-branch denoising module, utilizing both structural and evolutionary information. Additionally, we present a coupled conditional diffusion module that iteratively refines the evolution process by incorporating global and local conditions. This allows the model to incorporate more functional information than traditional antibody inverse folding models. (3) Empirical experiments demonstrate that our method improves the state-of-the-art methods in multiple antibody inverse folding tasks,  e.g., an 8.08\% AAR gains in long CDRH3 inverse folding task and an average of 7 cal/mol absolute $\Delta$$\Delta$G improvements in functionality optimization task, offering a fresh perspective on biomolecular generative models.

\section{Related work}
\label{gen_inst}
\textbf{Antibody Design}  Computational antibody design primarily follows two paths: conventional energy function optimization methods and machine learning approaches. Early antibody design methods were often limited to sequence similarity and energy function optimization \citep{lapidoth2015abdesign,adolf2018rosettaantibodydesign}. Recent success of machine learning approaches mainly falls into two directions: antibody sequence design and antigen-specific antibody sequence-structure co-design. 
The methods used for antibody sequence design mainly include language-based models \citep{ruffolo2021deciphering,olsen2022ablang,wang2023pre} and inverse folding models \citep{dreyer2023inverse,hoie2024antifold}. 
The other line focuses on antibody sequence-structure co-design mainly taking antibody-antigen complex as a graph, then using graph networks to extract features and predict the coordinates and residue type of antibody CDR \citep{jiniterative2021,kongMEAN,kong2023end,lin2024geoab, luo2022antigen,zhuabx,martinkus2024abdiffuser}. 
While these works are undoubtedly powerful, they often generate antibodies from scratch without incorporating explicit structure constraints, which can introduce challenges in designing functional antibodies \citep{zhou2024antigen}.
Instead, our method leverages the power of templates from a structure-informed perspective.

\textbf{Diffusion generative models} Diffusion models \citep{sohl2015deep,song2020denoising,ho2020denoising} are a class of generative models that have achieved impressive progress on a lot of generation tasks. Denoising diffusion probabilistic models (DDPMs) are a branch of diffusion models, which contain two Markov processes. The forward process perturbs the data into pure noise, and then learns to generate data by reversing the forward Markov process. Because of the diffusion model's flexibility and controllability, numerous works are focusing on employing retrieval-augmented methods to complement the diffusion framework for text-to-image generation \citep{sheyninknn}, image generation \citep{blattmann2022retrieval}, human motion generation \citep{zhang2023remodiffuse} and small molecule generation \citep{huanginteraction}.

\textbf{Retrieval augmented generative models}
Retrieval augmented generation technique was first proposed in the field of natural language processing to enhance the language models by introducing an additional database \citep{lewis2020retrieval,pmlr-v119-guu20a}, prompting the language models to generate more realistic and diverse results.
Subsequently, retrieval augmented generation (RAG) has conducted diverse explorations in large fields, including natural language processing \citep{zhang2023retrieve,gao2023retrieval,xuretrieval,yoranmaking,caffagni2024wiki} and computer vision \citep{long2022retrieval,blattmann2022retrieval,hu2023reveal,rao2023retrieval}. 

Recently, several studies have been proposed using retrieval techniques to enhance molecular generation. RetMol \citep{wang2023retrievalbased}  generates new molecules based on existing small molecules by retrieving a set of exemplar moleculars. IRDiff \citep{huanginteraction} enhances protein-specific molecular generation by using protein pockets to retrieve moleculars which interct with the pocket.
Although there has been some retrieval-based work in the field of protein design and discovery \citep{zhou2015rapid,aguilar2022fragment}, to the best of our knowledge, this is the first-of-its-kind retrieval-based generative framework for antibody design.

\section{Preliminaries and Notations}
\subsection{Notations}
\label{3.1}
Antibody consists of two heavy chains and two light chains. Each chain's tip has a complementary site that specifically binds to a unique epitope on the antigen. This site includes six complementary-determining regions (CDRs): CDR-H1, CDR-H2, and CDR-H3 on the heavy chain, and CDR-L1, CDR-L2, and CDR-L3 on the light chain \citep{presta1992antibody,al1997standard}.

Our work represents each single protein residue in terms of the residue type $
s_i\in \left\{ ACDEFGHIKLMNPQRSTVWY \right\}$, the coordinate $\evx_i\in \mathbb{R}^3$, and the orientation $\textbf{O}_i\in \text{SO}\left( 3 \right) $,  
where $i=1,..., N $ and N is the number of residues in the complex.  Concretely, assuming the CDR sequence to be generated includes $m$ amino acids and starts from position $a$, it can be denoted as 
$R=\left\{ s_j \mid j\in \{a+1,...,a+m\} \right\} $. Let $M$ be the length of the antibody, the antibody framework is defined as $ C_{ab}=\left\{ \left( s_i,x_j,\textbf{O}_j \right) \mid i\in \left\{ 1,..., M \right\} \backslash\{a+1,...,a+m \}, j\in \left\{1,..., M \right\} \right\} $. The antibody framework sequence is defined as $S_{ab} = \left\{s_i\mid i\in \left\{ 1,...,M \right\} \backslash\{a+1,...,a+m \}\right\}$. The antigen is defined as $ C_{ag}=\left\{ \left( s_i,x_i,\textbf{O}_i \right) \mid i\in \left\{ M+1,...,N \right\} \right\} $. 
The retrieved CDR-like fragments are defined as $\sA = \left\{A_i \mid i\in \left\{ 1,...,k \right\} \right\} $.
The goal of our framework is to extract the antibody CDR structure from the antibody framework complex $C_{ab}$, then input it into the retrieval module to retrieve $\sA$, and ultimately predict the distribution of $R$ through $C_{ag}$, $C_{ab}$ and $\sA$.
\subsection{Diffusion model for antibody design}
\label{3.2}
Due to diffusion models' excellent performance and controllability, there are now many diffusion-based works that have achieved notable results \citep{luo2022antigen,villegas-morcillo2023guiding,kulytė2024improving}. To be specific, they are denoising probabilistic diffusion models that transform the amino acid type $s$, the backbone $C\alpha$ atom coordinates $x$, and the amino acid orientation $\textbf{O}$ during the diffusion process. We focus on sequence, of which the forward process perturbs the data in the following ways \citep{multHoo}:
\begin{equation}
q\left(\mathrm{~s}_j^t \mid \mathrm{s}_j^{t-1}\right)=\operatorname{Multinomial}\left(\left(1-\beta^t\right) \cdot \operatorname{onehot}\left(\mathrm{s}_j^{t-1}\right)+\beta^t \cdot \frac{1}{20} \cdot \mathbf{1}\right)
\end{equation}
where $\beta^t$ is the noise schedule for the diffusion process, as \textit{t} approaches T, $\beta^t$ will approach 1, and the probability distribution will become closer to pure noise. \textbf{1} corresponds to a 20-dimensional all-one vector. 

To reverse the aforementioned forward process and denoise to generate CDR sequence, predictions need to be made by a neural network $F$(·)$[j]$, which takes the antibody-antigen context as condition:
\begin{equation}
    p\left( \text{s}_{j}^{t-1}\mid \mathcal{R}^t,\mathcal{C}_{ab}, \mathcal{C}_{ag} \right) = \operatorname{Multinomial}\left( F\left( \mathcal{R}^t,\mathcal{C}_{ab}, \mathcal{C}_{ag} \right) \left[ j \right] \right) 
\end{equation}

As an example, this work uses Diffab \citep{luo2022antigen} as the backbone for the generative model to conduct retrieval augmented generation. Note that the proposed retrieval system is generative model agnostic, and the developed modules can be integrated with any diffusion generative model.

\section{Methods}
\begin{figure}[t]
        \centering
        \includegraphics[width=1\textwidth]{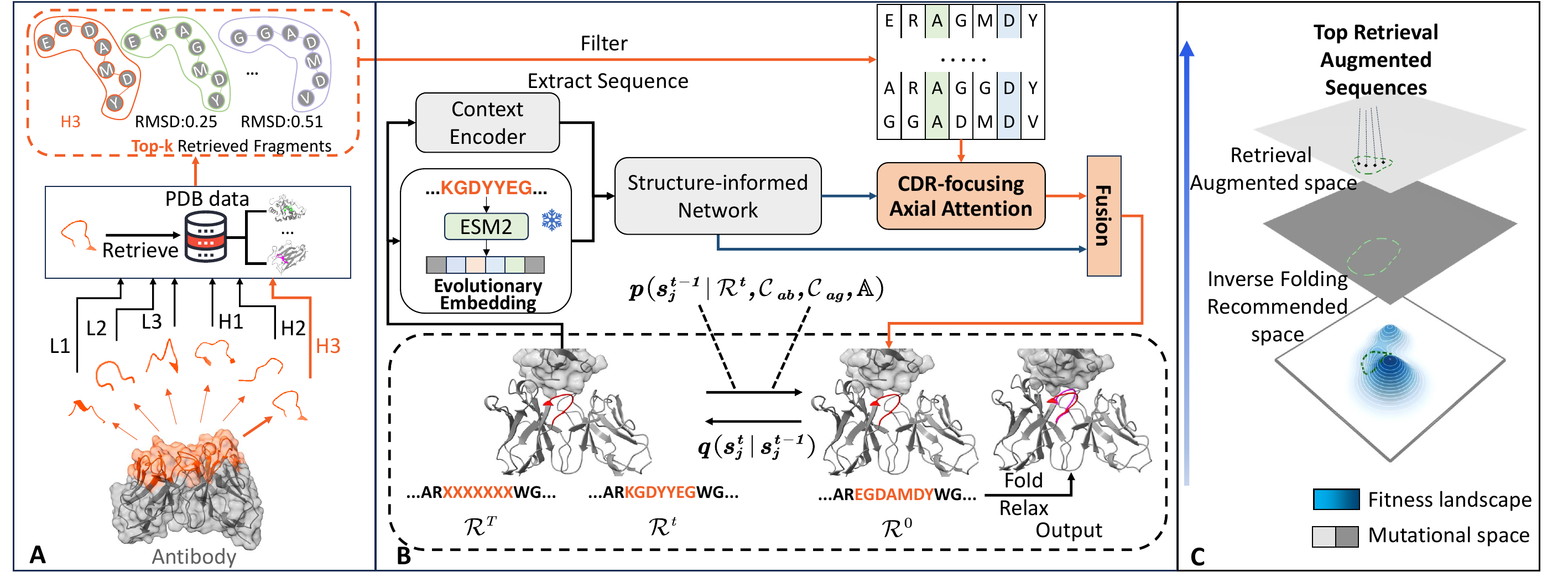}
        \vspace{-3mm}
        \caption{The overall architecture of the proposed RADAb framework. (A) Structural retrieval process, the CDR backbone is input into MASTER and the output is a set of ranked CDR-like fragments. (B) Diffusion process and denoising network which takes antibody-antigen context and retrieved evolutionary information as conditions. The structure is fixed during diffusion process. (C) Our method restricts the antibody to a small region through fixed structural constraints and retrieval-augmented constraints (functional constraints) to achieve higher fitness.}
        \label{main_figure}
        \vspace{-4mm}
\end{figure}

We propose RADAb (as demonstrated in Figure \ref{main_figure}), a novel structure-informed retrieval-augmented diffusion framework for antibody sequence design and optimization. The model uses a structural retrieval algorithm to search for antibody homologous structures and take their sequences as conditional inputs for the diffusion model to provide homologous patterns and evolutionary information. 

\subsection{Structural retrieval of CDR fragments}
\label{4.1}
The structure of a protein is determined by its sequence, and protein sequences that can fold into similar structures exhibit similar properties. These structurally similar protein sequences contain rich evolutionary information. Based on this, we perform retrieval in the PDB database using CDR structures, aiming to obtain fragments that are similar to the real CDR and have homologous sequences, with the expectation that they possess similar functions. 

To balance the quality of results and the retrieval speed, we use MASTER \citep{zhou2015rapid} for the search. MASTER uses the root-mean-square deviation (RMSD) of backbone atoms as a similarity measure. It queries structural fragments composed of one or more non-contiguous segments and can find all matching fragments from the database within a given RMSD threshold. This allows for fast and accurate searches in the PDB database for protein motifs. Note that MASTER can utilize only the backbone information without any leakage of sequence data during the search process. The retrieval procedure is described in Algorithm \ref{alg_1} and detailed in Appendix \ref{app_a.3}. 

For the retrieved results, we use the RMSD with the real backbone structure as a score to rank them and filter out the input CDR fragment. For ease of use, we further constructed a CDR-like fragments database (detailed in Appendix \ref{appa.4}). Additionally, to enable the model to learn richer evolutionary information, we filter out identical CDR-like sequences during the training phase. However, to improve the quality of the model's generation, we do not perform similar filters during generation.
\begin{algorithm}[h]
    \caption{Structural Retrieval Algorithm Overview}
    \label{alg_1}
    \begin{algorithmic}[1]
        \State {\bfseries Input:} Coordinates set $\mathcal{X} = \{x_{k}\mid k\in  \left\{ 1,...,m \right\}\}$
        \State {\bfseries Input:} Structure database with $P$ structures $\sT=\left\{\tau_{i} \mid i\in \left\{ 1,...,P \right\} \right\}$
        \State Initialize CDR-like fragments set: $\sA \leftarrow \emptyset$
        \State Initialize structure residues set: $\mathcal{C} \leftarrow \emptyset$
        \State Initialize threshold maxA(), maxB(), maxC()
        \For{$i = 1$ {\bfseries to} $P$}
            \State $\mathcal{C} \leftarrow$ all residues in $\tau_i$
            \For{each residue $j$ in $\mathcal{C}$}
                \State $r \leftarrow RMSD(\mathcal{X}, j)$
                \If{$r > \text{maxA}(\mathcal{X})$}
                    \State eliminate $j$ from the list $\mathcal{C}$
                    \State continue
                \EndIf
                \If{$(r > \text{maxB}(\mathcal{X}))$ {\bfseries OR} $(cRMSD(\mathcal{X}) > \text{maxC}(\mathcal{X}))$}
                    \State continue
                \EndIf
                \State $A \leftarrow$ $J$ 
                \State insert match $A$ into $\sA$
            \EndFor
        \EndFor
        \State {\bfseries return} $\sA$
    \end{algorithmic}
\end{algorithm} 

\subsection{Model Architectures}
\label{4.2}
The model takes the antigen-antibody complex's structure and sequence context, along with the sequences of the CDR-like fragments, as conditional inputs to iteratively denoise. 
The first branch of the model learns the global context information of the complex, while another branch takes the local homologous information of CDR-like fragments as input, aiming to learn the functional similarity and evolutionary information of residues with similar structure. The two branches are combined to generate the antibody CDR sequence jointly. 

\subsubsection{Global geometry context information branch}
\textbf{Context encoder} 
A protein is formed by the connection of multiple residues. The features of a single residue mainly include the residue type, backbone atom coordinates, and backbone dihedral angles. The features of each pair of residues mainly include the types of both residues, sequential relative position, spatial distance, and pairwise backbone dihedrals. These features are concatenated and then input into two separate MLPs. The output is denoted as $z_i$ and $y_{ij}$.

\textbf{Evolutionary encoder}
Recent advances have shown the structure-informed protein language model (PLM) is an excellent tool for creating protein sequence embeddings and providing evolutionary information \citep{zheng2023structure,shanker2024unsupervised}. Thus, we take ESM2 \citep{lin2023evolutionary} as an antibody sequence encoder, aiming to capture the evolutionary relations of antibody residues. The state of antibody sequence with CDR at timestep $t$ is fed into it and output is defined as $e^t$.
 
\textbf{Structure-informed network}
The above encoding is used as conditional input to the Structure-informed network. They, along with the CDR sequence and structural state at the current time step, will be input into a stack of Invariant Point Attention (IPA) \citep{jumper2021highly} layers, and jointly transform into a hidden representation $h_i$. 
Subsequently, the hidden representation $h_i$ is transformed by an MLP to obtain the probability representation $r_{\text{global}}$ of the amino acid type at each CDR site. This probability representation is then input to the local CDR-focused branch.

\subsubsection{Local CDR-focused information branch} 
\textbf{Post-processing of CDR-like fragments}
We first remove the CDR portions from the antibody sequences to obtain the antibody framework sequences. Then, we fill these fragments' short sequences into the antibody framework sequences, thereby constructing a CDR-like sequence matrix $\rmE$. 

\textbf{CDR-focused Axial Attention}
The local CDR-like branch is constituted of a stack of axial attention layers, referred to as CDR-focused Axial Attention. Given that the CDR-like fragments exhibit structures similar to actual CDRs, we employed a tied row attention mechanism used in MSATransformer \citep{rao2021msa} to leverage these retrieval results. In the standard axial attention \citep{ho2019axial} mechanism, each row and column are considered independently. However, in MSA (Multiple Sequence Alignment), each sequence exhibits relatively similar structural features. Our matrix format is well-suited for adopting a tied row attention mechanism to fully utilize the structural similarity. When calculating the attention scores for each row, this mechanism simultaneously considers the scores of other rows. This approach not only leverages the structural similarity but also reduces memory usage.

The input to CDR-focused Axial Attention is a pseudo-MSA matrix $\rmP$ in equation \ref{equ:P}. The first row of this matrix is initially filled with the antigen-antibody framework sequence, with the CDR region populated by the noisy sequence $R^t_{g}$ (sampled by $r_{\text{global}}$) at the current time step $t$. From the second row to the k-th row (where k is chosen to be 16, meaning the top 15 retrieved CDR-like sequences are used as conditional input), the rows are filled with the CDR-like sequence matrix $\rmE$. The constructed matrix $\rmP$ is then input to CDR-focused Axial Attention to create the homologous embedding and calculate the probability representation $r_{\text{local}}$ (equation \ref{equ:MSA}).
\begin{equation}
\label{equ:P}
    \rmP = \operatorname{concat}\left(\left(S_{ab}\cup R^t_{g} \right), \rmE\right)
\end{equation}

\vspace{-4mm}
\begin{equation}
\label{equ:MSA}
\begin{aligned}
    r_{\text{local}}[\cdot , j] &= G_{\text{col}}\left(\rmP_{\cdot,j}, t \right) \text{ for all } j \in \text{col}, \\
    r_{\text{local}}[i, \cdot] &= G_{\text{tiedrow}} \left(\rmP_{i,\cdot}, t \right) \text{ for all } i \in \text{row}
\end{aligned}
\end{equation}

The row self-attention is computed to capture the internal relationships within the antibody-antigen sequences, while the column self-attention is computed to capture the relationships between the CDR residues and the CDR-like residues.

\textbf{Skip connection for information fusion} 
Although the probability distribution of the CDR region created by the antigen-antibody context features has already been fed into the network, to prevent the loss of antigen-antibody context information during forward propagation, the embedding  $r_{\text{local}}$ and $r_{\text{global}}$ are added by a skip connection module \citep{he2016deep}, then execute \textit{softmax} to obtain the final probability distribution.

\subsection{Model training and inference}
\label{4.3}
\textbf{The overall training objective}
The training objective is to minimize the probability distributions predicted by the network under two conditions at each time step and the true posterior distribution at the same time step. Therefore, we choose KL divergence between the two distributions at each residue in the CDR region as the training loss function,
\begin{equation}
    L_{\text{type\,\,}}^{t}=\mathbb{E}_{\mathcal{R}^t\sim p}\left[ \frac{1}{m}\sum_j{D}_{\text{KL}}\left( q\left( \text{s}_{j}^{t-1}\mid \text{s}_{j}^{t},\text{s}_{j}^{0} \right) ||p\left( \text{s}_{j}^{t-1}\mid \mathcal{R}^t,\mathcal{C}_{ab}, \mathcal{C}_{ag},\sA \right) \right) \right] 
\end{equation}
The training objective of the whole diffusion process is:
\begin{equation}
    L=E_{t\sim \,\,\text{Uniform\,\,}\left( 1...T \right)} L_{\text{type\,\,}}^{t}
\end{equation}

\textbf{Conditional reverse diffusion process}
We employ DDPM to generate sequences. The model starts from time step T, initializing each site of the antibody CDR region as a uniform distribution. Then, through the frozen ESM encoder $E$(·), learned global context network $F$(·)$[j]$ and the local CDR-focused network $G$(·)$[j]$, they predict the noise distribution at each time step jointly and denoise step-by-step:
\begin{equation}
    e^t = E(S_{ab} \cup R^t)
\end{equation}
\vspace{-2mm}
\begin{equation}
    \begin{aligned}
    p\left( s_{j}^{t-1} \mid \mathcal{R}^t, \mathcal{C}_{ab}, \mathcal{C}_{ag}, \sA \right) = \operatorname{Multinomial}\big[ F\left( \mathcal{R}^t, \mathcal{C}_{ab}, \mathcal{C}_{ag}, e^t \right) \\
    + G \left( F\left( \mathcal{R}^t, \mathcal{C}_{ab}, \mathcal{C}_{ag}, e^t \right), \sA \right) \big] \left[ j \right]
    \end{aligned}
\end{equation}
During sampling, we remove the CDR region sequences from the antibody structures and fill them with noisy sequence sampled from the uniform distribution. The retrieval process uses the structure of the CDR region as input and outputs a set of CDR-like fragments. Subsequently, this set of CDR-like fragments is fed into the retrieval-augmented diffusion model, serving as a condition along with the antigen-antibody framework context to guide the model in step-by-step denoising and generating the CDR sequences.

\section{Experiments}

To evaluate the performance of our model's generation, we utilize two tasks: antibody CDR sequence inverse folding (Section  \ref{5.1}) and antibody optimization based on sequence design (Section  \ref{5.2}), to compare with the baselines. Additionally, we conducted ablation experiments and further analysis to demonstrate the effectiveness of the retrieval-augmented method (Section \ref{5.3}).

The dataset for training the model is obtained from the SAbDab and our established CDR-like fragments dataset. Following the previous work \citep{luo2022antigen}, we first eliminated structures with a resolution lower than 4Å and removed antibodies that target non-protein antigens. Chothia \citep{chothia1987canonical} in ANARCI \citep{dunbar2016anarci} is used for renumbering antibody residues. We clustered the SADab datasets based on 50\% sequence similarity in the CDR-H3 region, and chose 50 PDB files comprising 63 antibody-antigen complex structures as the test set. To ensure distinct training and test sets, we removed structures from the training set that were part of the same clusters as those in the test set. 

\subsection{Antibody CDR Sequence inverse folding}
\label{5.1}
\textbf{Baselines} For traditional methods, we simulated a method of grafting using CDR-like data in the process of rational antibody design. Specifically, we directly graft the retrieved top-1 CDR-like fragment onto the antibody framework, termed \textbf{Grafting}.
For deep learning methods, we selected a series of state-of-the-art protein inverse folding models for comparison with our work, including \textbf{ProteinMPNN} \citep{dauparas2022robust}, a model that utilizes message passing neural network to design sequences with a fixed protein backbone;\textbf{ ESM-IF} \citep{hsu2022learning}, a protein inverse folding model that trained on millions of predicted structures; \textbf{Diffab-fix} \citep{luo2022antigen}, which can fix the backbone structure and iteratively generate candidate sequences from pure noise in sequence space using diffusion; \textbf{AbMPNN} \citep{dreyer2023inverse}, a model fine-tuned ProteinMPNN on antibody sequence and structure. Because it is not open-sourced, we evaluate it on its own test set. For more baseline details, please refer to Appendix \ref{appa.5}.
\begin{table}[t]
\vspace{-3mm}
\caption{Results of sequence design on SAbDab dataset}
\begin{center}
\label{tab1}
\arrayrulecolor{black}
\resizebox{.996\textwidth}{!}{
\begin{adjustbox}{width=\textwidth}%
\fontsize{16}{16}\selectfont
\begin{tabular}{cccccccccc}
\toprule
\multirow{2}{*}{Method} & \multicolumn{3}{c}{CDR-H1}      & \multicolumn{3}{c}{CDR-H2}      & \multicolumn{3}{c}{CDR-H3}      \\
                        & AAR(\%) $\uparrow$ & scRMSD $\downarrow$ & Plausibility $\uparrow$ & AAR(\%) $\uparrow$ & scRMSD $\downarrow$ & plausibility $\uparrow$ & AAR(\%) $\uparrow$ & scRMSD $\downarrow$ & plausibility $\uparrow$ \\ 
                     
\midrule
Grafting   &58.05 &0.83 &-0.597 &31.46 &0.79 &-0.619 & 19.63 &3.20 & -0.591\\
\midrule
ProteinMPNN             & 58.58 & 0.64 & -0.603 & 53.18 & 0.61 & -0.568 & 41.77 & 2.27 & -0.605 \\
ESM-IF1                 & 53.80 & 0.66 & -0.610 & 46.66 & 0.63 & -0.589 & 29.82 & 2.59 & -0.607 \\
Diffab-fix                  & 74.93 & 0.66 & -0.512 & 65.41 & 0.59 & -0.532 & 49.17 & 2.24 & -0.541 \\
AbMPNN*                 & 72.83 & 1.09 & -0.664 & 65.33 & 0.93 & -0.677 & 52.99 & 2.80 & -0.675 \\
\midrule
\textbf{RADAb}          & \textbf{76.57} & \textbf{0.61} & \textbf{-0.505} & \textbf{66.16} & \textbf{0.57} & \textbf{-0.530} & \textbf{57.02} & \textbf{2.23} & \textbf{-0.530} \\
\midrule

\midrule
\multirow{2}{*}{Method} & \multicolumn{3}{c}{CDR-L1}      & \multicolumn{3}{c}{CDR-L2}      & \multicolumn{3}{c}{CDR-L3}      \\
                        & AAR(\%) $\uparrow$ & scRMSD $\downarrow$ & Plausibility $\uparrow$ & AAR(\%) $\uparrow$ & scRMSD $\downarrow$ & plausibility $\uparrow$ & AAR(\%) $\uparrow$ & scRMSD $\downarrow$ & plausibility $\uparrow$ \\ 
\midrule
Grafting     &68.53 &0.85 & -0.506 & 43.19 & 0.52&-0.573 &43.61 &1.08&-0.395 \\
\midrule
ProteinMPNN             & 45.60 & 0.59 & -0.612 & 46.78 & 0.46 & -0.527 & 47.21 & 0.98 & -0.543 \\
ESM-IF1                 & 40.97 & 0.61 & -0.650 & 43.40 & 0.43 & -0.542 & 38.93 & 0.92 & -0.569 \\
Diffab-fix                  & 79.78 & 0.56 & -0.386 & 81.19 & 0.44 & -0.398 & 67.97 & 0.88 & -0.414 \\
AbMPNN*                 & 75.06 & 0.73 & -0.543 & 71.63 & 0.56 & -0.528 & 64.51 & 0.91 & -0.544 \\
\midrule
\textbf{RADAb}          & \textbf{83.72} & \textbf{0.54} & \textbf{-0.379} & \textbf{84.58} & \textbf{0.44} & \textbf{-0.384} & \textbf{73.11} & \textbf{0.87} & \textbf{-0.384} \\
\bottomrule %
\end{tabular}
\end{adjustbox}
}
\end{center}
\vspace{-5mm}
\end{table} 

\textbf{Metrics} To evaluate the accuracy and rationality of the sequences generated by the model, we selected the following three popular evaluation metrics:  (1) Amino Acid Recovery (AAR,$\%$): AAR refers to the ratio of positions where the designed sequence and the true CDR sequence have the same amino acid; (2) Self-consistency RMSD (scRMSD, Å ): To calculate scRMSD, we refold the antibody sequences generated by the model using ABodyBuilder2 \citep{abanades2023immunebuilder}. Then, we align the refolded antibody framework with the original antibody and compute the RMSD of the $C_\alpha$ atoms in the CDR region. (3) Plausibility: We use pseudo-log-likelihood in an antibody language model, AntiBERTy \citep{ruffolo2021deciphering} to calculate plausibility of the generated sequence.

\textbf{Results} As shown in Table \ref{tab1}, RADAb outperforms state-of-the-art methods in each metric and at each CDR region. In particular, in the highly variable and specific CDR-H3 region \citep{shirai1999h3,raybould2019five}, our method achieved a great improvement in AAR compared to best-performing methods Diffab-fix and AbMPNN. The evaluation results indicate that the retrieval-augmented method, by introducing structurally similar homologous sequences, has improved the accuracy, consistency, and rationality of the model's generation. 

In addition, CDR-H3 exhibits significant variability in length, sequence, and structure. Typically, deep learning models show decreased performance when generating longer CDR-H3 sequences \citep{luo2022antigen,hoie2024antifold}. Therefore, we selected a subset of the test set with CDR-H3 lengths longer than 14 to evaluate the generation performance. As shown in Table \ref{long_CDR3},  while the generation performance of all methods declines to some extent, our method demonstrates consistency and significantly outperforms the others, with a larger margin of improvement.
\begin{table*}[t]
\vspace{-3mm}
    \begin{minipage}[t]{0.47\textwidth}
        \centering
        \caption{Results of long CDRH3 sequence design performance.}
        \begin{tabular}{@{}>{\arraybackslash}p{0.25\linewidth} >{\centering\arraybackslash}p{0.19\linewidth} >{\centering\arraybackslash}p{0.19\linewidth}>{\centering\arraybackslash}p{0.24\linewidth}@{}}
            \toprule
            Method & AAR(\%) & scRMSD & plausibility \\
            \midrule
            Grafting &7.79& 4.05 &-0.785 \\
            \midrule
            ProteinMPNN & 46.63 & 2.71 & -0.820 \\
            ESM-IF1 & 30.01 & 2.86 & -0.845 \\
            Diffab-fix & 42.26 & 3.02 & \textbf{-0.740} \\
            AbMPNN* & 43.27 & 4.39 &  -1.012 \\
            \midrule
            \textbf{RADAb} & \textbf{51.35} & \textbf{2.52} & -0.747 \\
            \bottomrule
        \end{tabular}
        \label{long_CDR3}
    \end{minipage}
    \hfill
    \begin{minipage}[t]{0.48\textwidth}
        \centering
        \caption{Results of binding energy optimization based on antibody sequence design.}
        \begin{tabular}{@{}>{\arraybackslash}p{0.23\linewidth} >{\centering\arraybackslash}p{0.18\linewidth} >{\centering\arraybackslash}p{0.17\linewidth} >{\centering\arraybackslash}p{0.23\linewidth} 
        @{}}
            \toprule
             \multirow{2}*{Method} & \multirow{2}*{$\Delta$$\Delta$G$\downarrow$}  & $\Delta$$\Delta$G-seq $\downarrow$ & IMP-seq(\%) $\uparrow$  \\
            \midrule
            Grafting &135.17 &40.22 &32.69\\
            \midrule
            ProteinMPNN & 127.14 & 24.72 & 35.51 \\
            ESM-IF1 & 162.09 & 42.28 & 33.33 \\
            Diffab-fix & 116.36 & 14.05 & 34.52\\
            \midrule
            \textbf{RADAb} & \textbf{109.16} & \textbf{7.06} & \textbf{37.30}\\
            \bottomrule
        \end{tabular}
        \label{energy}
    \end{minipage}
\end{table*}
\vspace{-2mm}
\begin{figure}[t]
        \centering
        \includegraphics[width=1\textwidth]{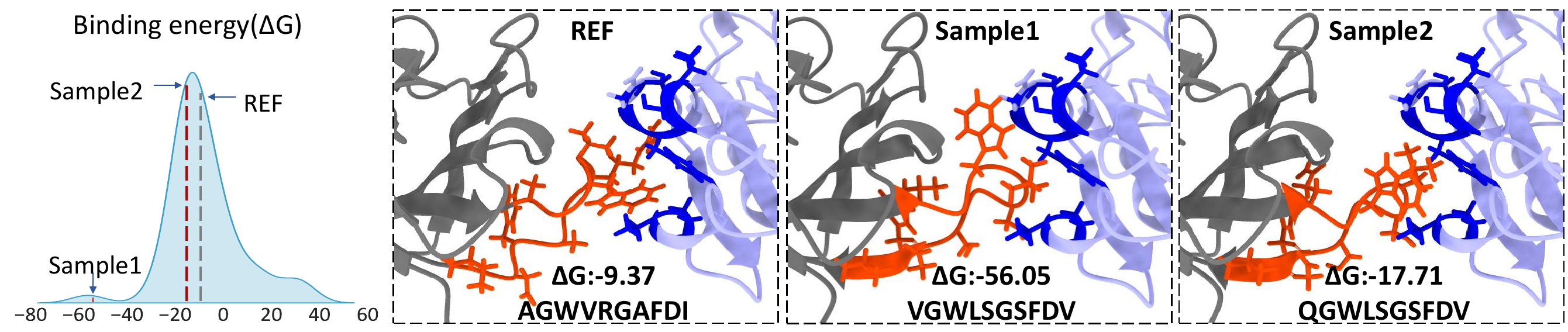}
        \vspace{-6mm}
        \caption{Left: Distribution of the samples' interface energy. Right: Generated CDR-H3 samples and the original structure of PDB: 7d6i antigen-antibody complex. The gray part represents the antibody framework, the red part represents the CDR, and the blue part represents the antigen (with the darker shade indicating the antigen epitope).}
        \label{casestudy}
        \vspace{-4mm}
\end{figure}
\subsection{Antibody functionality optimization}
\label{5.2}
In this section, we focus on the evolution of antibody sequences and evaluate whether the structure of the evolved sequence has greater functionality compared to the structure of the folded original sequence. To this end, we first fold the designed CDR-H3 sequences with framework sequences and the original real antibody sequences into complete protein structures using ABodyBuilder2. Then, we use \textit{FastRelax} and \textit{InterfaceAnalyzer} in PyRosetta \citep{alford2017rosetta} to relax the structure and calculate the binding energy $\Delta$G of the antibody-antigen complex.

\textbf{Metrics} 
We use various metrics to evaluate the efficacy and functionality of our designed antibodies: (1) $\Delta$$\Delta$G: This metric represents the difference in binding energy between the complex with the designed CDR folded into the structure and the original complex binding energy. (2) $\Delta$$\Delta$G-seq: This metric measures the difference in binding energy between the complex with the designed CDR sequence folded into the structure and the binding energy of the original antibody sequence folded into the structure. It aims to eliminate errors introduced by the folding tool, allowing for a direct comparison of sequence functionality. (3) IMP-seq: This metric indicates the percentage of designed CDR sequences folded into the structure with a lower (better) binding energy than the original antibody sequences folded into the structure.

\textbf{Results}
As shown in Table \ref{energy}, after folding and relaxing, the antibody sequences we designed show a significant decrease in binding energy compared to other methods, and 37.3\% of them have better binding energy than those folded from the original antibody sequences.

To further demonstrate the optimization of antibody sequence functionality by RADAb, we select a specific antigen-antibody complex from the test set (A neutralizing MAb targeting the receptor-binding domain of SARS-CoV-2, PDB: 7d6i). We generate 50 sequences for CDR-H3 and calculate the binding energy $\Delta$G of the folded structures. Among these, 68\% of the samples exhibited lower $\Delta$G compared to the original complex. As shown in Figure \ref{casestudy}, we select two samples as examples. Although they do not achieve the highest AAR, they demonstrate better binding affinity compared to the native structure.

\subsection{Analysis}
\label{5.3}
\textbf{Ablation} We conducted a series of ablation experiments for CDR-H3 following the settings described in Section \ref{5.1} to validate the effectiveness and relative contributions of the additional conditions and data we introduced. The specific objectives are: (1) to verify the effectiveness of the retrieval augment module; (2) to assess the validity of the retrieved data; and (3) to evaluate the effectiveness of the evolutionary embedding. 
\begin{figure}[t]
    \vspace{-7mm}
    \centering
    \begin{minipage}{0.5\textwidth} %
        \centering
        \begin{table}[H]
            \caption{Ablation Study. \textit{G} represents using Ground truth as retrieval results, \textit{R} represents the Retrieval-augment mechanism, and \textit{E} represents Evolutionary embedding mechanism.
            }
            \centering
            \small %
            \renewcommand{\arraystretch}{0.8} %
            \begin{tabular}{cccccc} %
            \toprule
            \multicolumn{3}{c}{Ablation}& \multirow{2}*{AAR(\%)} & \multirow{2}*{scRMSD} & \multirow{2}*{Plausibility} \\
            \textit{G}& \textit{R}& \textit{E}& & &\\ 
            \midrule
            \color{gray}\CheckmarkBold & \color{gray}\CheckmarkBold & \color{gray}\CheckmarkBold & \multicolumn{1}{c}{\color{gray}70.56} & \multicolumn{1}{c}{\color{gray}2.13} & \multicolumn{1}{c}{\color{gray}-0.534} \\
            \midrule
            \XSolidBrush & \XSolidBrush & \CheckmarkBold & 51.36 & {2.23} & {-0.538} \\
            \XSolidBrush & \CheckmarkBold & \XSolidBrush & {52.15} & {2.39} & {\textbf{-0.529}} \\
            \XSolidBrush & \XSolidBrush & \XSolidBrush & {49.17} & {2.24} & {-0.541} \\
            \midrule
            \XSolidBrush & \CheckmarkBold & \CheckmarkBold & {\textbf{57.02}} & {\textbf{2.23}} & {-0.530} \\
            \bottomrule
            \end{tabular}
            \label{ablation}
            \vspace{-3mm}
        \end{table}
    \end{minipage}
    \hfill
    \begin{minipage}{0.47\textwidth}
        \vspace{8mm}
        \centering
        \includegraphics[width=1\textwidth]{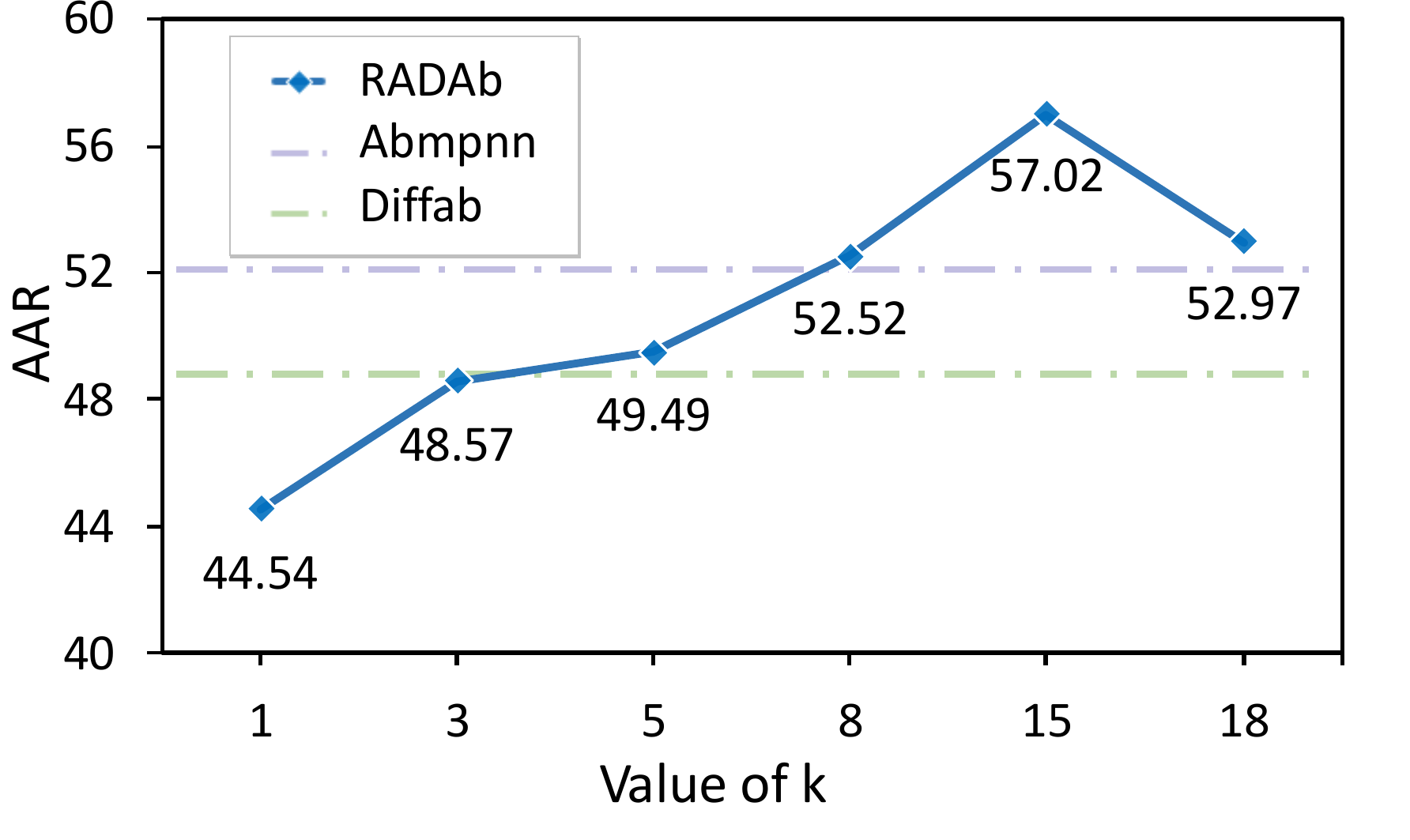}
        \vspace{-6mm}
        \caption{The effect of the value of CDR-like fragments $k$ on model's CDRH3 performance.}
        \label{num-of-k}
        \vspace{-3mm}
    \end{minipage}
\end{figure}

As shown in Table \ref{ablation}, We demonstrated the retrieval augment module's effectiveness by inputting the CDR sequence's ground truth into this module. 
We also removed the retrieval augmentation mechanism and the evolutionary embedding mechanism respectively to validate their effectiveness. The experimental results show that both the retrieval augmentation module and the evolutionary embedding module individually improve performance, and using them together maximizes the model's performance.

\textbf{Effect of retrieval dataset} To further analyze the benefits brought by the retrieval mechanism and retrieved motifs, we conducted a series of comparative experiments on the value $k$ of CDR-like fragments selected as conditions in the diffusion network, as shown in Figure \ref{num-of-k}. 
We found that when the value of $k$ is low, it brings negative benefits to the model, which may be due to overfitting. As the value of $k$ increases, the model performance also gradually improves. When $k$ equals 15, the model achieves the best performance. But when $k$ exceeds 15, the additional information instead introduces noise to the model, leading to performance degradation.

\section{Conclusion and Future works}
In this work, we propose a retrieval-augmented diffusion generative model RADAb for antibody sequence design. This model leverages global geometric information and local template information, incorporating these conditions into the diffusion process to enhance antibody sequence design and optimization. Experimental results demonstrate that RADAb achieves state-of-the-art performance across multiple tasks. The main limitation of this work is that it has not yet been fully validated in wet lab experiments, which will be one of the major tasks in the future. Since we have proposed a comparatively general retrieval method and retrieval-augment framework, another major future task is to extend the model to the design of various protein motifs.

\subsubsection*{Reproducibility statement}
The code is avalibale at \href{https://github.com/GENTEL-lab/RADAb}{https://github.com/GENTEL-lab/RADAb}

\subsection*{Acknowledgments}
This study has been supported by the National Natural Science Foundation of China [62041209], Natural Science Foundation of Shanghai [24ZR1440600], the Science and Technology Commission of Shanghai Municipality [24510714300].

\normalem
\bibliography{iclr2025_conference}

@article{ewert2004stability,
  title={Stability improvement of antibodies for extracellular and intracellular applications: CDR grafting to stable frameworks and structure-based framework engineering},
  author={Ewert, Stefan and Honegger, Annemarie and Pl{\"u}ckthun, Andreas},
  journal={Methods},
  volume={34},
  number={2},
  pages={184--199},
  year={2004},
  publisher={Elsevier}
}

@article{jones1986replacing,
  title={Replacing the complementarity-determining regions in a human antibody with those from a mouse},
  author={Jones, Peter T and Dear, Paul H and Foote, Jefferson and Neuberger, Michael S and Winter, Greg},
  journal={Nature},
  volume={321},
  number={6069},
  pages={522--525},
  year={1986},
  publisher={Nature Publishing Group UK London}
}

@article{xu2000diversity,
  title={Diversity in the CDR3 region of VH is sufficient for most antibody specificities},
  author={Xu, John L and Davis, Mark M},
  journal={Immunity},
  volume={13},
  number={1},
  pages={37--45},
  year={2000},
  publisher={Elsevier}
}

@article{akbar2021compact,
  title={A compact vocabulary of paratope-epitope interactions enables predictability of antibody-antigen binding},
  author={Akbar, Rahmad and Robert, Philippe A and Pavlovi{\'c}, Milena and Jeliazkov, Jeliazko R and Snapkov, Igor and Slabodkin, Andrei and Weber, C{\'e}dric R and Scheffer, Lonneke and Miho, Enkelejda and Haff, Ingrid Hob{\ae}k and others},
  journal={Cell Reports},
  volume={34},
  number={11},
  year={2021},
  publisher={Elsevier}
}

@article{sormanni2015rational,
  title={Rational design of antibodies targeting specific epitopes within intrinsically disordered proteins},
  author={Sormanni, Pietro and Aprile, Francesco A and Vendruscolo, Michele},
  journal={Proceedings of the National Academy of Sciences},
  volume={112},
  number={32},
  pages={9902--9907},
  year={2015},
  publisher={National Acad Sciences}
}

@article{lapidoth2015abdesign,
  title={Abdesign: A n algorithm for combinatorial backbone design guided by natural conformations and sequences},
  author={Lapidoth, Gideon D and Baran, Dror and Pszolla, Gabriele M and Norn, Christoffer and Alon, Assaf and Tyka, Michael D and Fleishman, Sarel J},
  journal={Proteins: Structure, Function, and Bioinformatics},
  volume={83},
  number={8},
  pages={1385--1406},
  year={2015},
  publisher={Wiley Online Library}
}

@article{olsen2022ablang,
  title={AbLang: an antibody language model for completing antibody sequences},
  author={Olsen, Tobias H and Moal, Iain H and Deane, Charlotte M},
  journal={Bioinformatics Advances},
  volume={2},
  number={1},
  pages={vbac046},
  year={2022},
  publisher={Oxford University Press}
}

@inproceedings{wang2023pre,
  title={On pre-training language model for antibody},
  author={Wang, Danqing and Fei, YE and Zhou, Hao},
  booktitle={The eleventh international conference on learning representations},
  year={2023}
}

@article{ruffolo2021deciphering,
  title={Deciphering antibody affinity maturation with language models and weakly supervised learning},
  author={Ruffolo, Jeffrey A and Gray, Jeffrey J and Sulam, Jeremias},
  journal={arXiv preprint arXiv:2112.07782},
  year={2021}
}

@article{hoie2024antifold,
  title={AntiFold: Improved antibody structure-based design using inverse folding},
  author={H{\o}ie, Magnus Haraldson and Hummer, Alissa and Olsen, Tobias H and Aguilar-Sanjuan, Broncio and Nielsen, Morten and Deane, Charlotte M},
  journal={arXiv preprint arXiv:2405.03370},
  year={2024}
}

@article{dreyer2023inverse,
  title={Inverse folding for antibody sequence design using deep learning},
  author={Dreyer, Fr{\'e}d{\'e}ric A and Cutting, Daniel and Schneider, Constantin and Kenlay, Henry and Deane, Charlotte M},
  journal={arXiv preprint arXiv:2310.19513},
  year={2023}
}

@inproceedings{jiniterative2021,
  title={Iterative Refinement Graph Neural Network for Antibody Sequence-Structure Co-design},
  author={Jin, Wengong and Wohlwend, Jeremy and Barzilay, Regina and Jaakkola, Tommi S},
  booktitle={International Conference on Learning Representations},
  year={2021}
}

@inproceedings{kongMEAN,
  title={Conditional Antibody Design as 3D Equivariant Graph Translation},
  author={Kong, Xiangzhe and Huang, Wenbing and Liu, Yang},
  booktitle={The Eleventh International Conference on Learning Representations},
  year={2022}
}

@inproceedings{kong2023end,
  title={End-to-end full-atom antibody design},
  author={Kong, Xiangzhe and Huang, Wenbing and Liu, Yang},
  booktitle={Proceedings of the 40th International Conference on Machine Learning},
  pages={17409--17429},
  year={2023}
}

@article{lin2024geoab,
  title={GeoAB: Towards Realistic Antibody Design and Reliable Affinity Maturation},
  author={Lin, Haitao and Wu, Lirong and Yufei, Huang and Liu, Yunfan and Zhang, Odin and Zhou, Yuanqing and Sun, Rui and Li, Stan Z},
  journal={bioRxiv},
  pages={2024--05},
  year={2024},
  publisher={Cold Spring Harbor Laboratory}
}

@inproceedings{zhuabx,
  title={Antibody Design Using a Score-based Diffusion Model Guided by Evolutionary, Physical and Geometric Constraints},
  author={Zhu, Tian and Ren, Milong and Zhang, Haicang},
  booktitle={Forty-first International Conference on Machine Learning},
  year={2024}
}

@article{luo2022antigen,
  title={Antigen-specific antibody design and optimization with diffusion-based generative models for protein structures},
  author={Luo, Shitong and Su, Yufeng and Peng, Xingang and Wang, Sheng and Peng, Jian and Ma, Jianzhu},
  journal={Advances in Neural Information Processing Systems},
  volume={35},
  pages={9754--9767},
  year={2022}
}

@article{martinkus2024abdiffuser,
  title={AbDiffuser: full-atom generation of in-vitro functioning antibodies},
  author={Martinkus, Karolis and Ludwiczak, Jan and Liang, Wei-Ching and Lafrance-Vanasse, Julien and Hotzel, Isidro and Rajpal, Arvind and Wu, Yan and Cho, Kyunghyun and Bonneau, Richard and Gligorijevic, Vladimir and others},
  journal={Advances in Neural Information Processing Systems},
  volume={36},
  year={2024}
}

@article{lewis2020retrieval,
  title={Retrieval-augmented generation for knowledge-intensive nlp tasks},
  author={Lewis, Patrick and Perez, Ethan and Piktus, Aleksandra and Petroni, Fabio and Karpukhin, Vladimir and Goyal, Naman and K{\"u}ttler, Heinrich and Lewis, Mike and Yih, Wen-tau and Rockt{\"a}schel, Tim and others},
  journal={Advances in Neural Information Processing Systems},
  volume={33},
  pages={9459--9474},
  year={2020}
}

@InProceedings{pmlr-v119-guu20a,
  title = 	 {Retrieval Augmented Language Model Pre-Training},
  author =       {Guu, Kelvin and Lee, Kenton and Tung, Zora and Pasupat, Panupong and Chang, Mingwei},
  booktitle = 	 {Proceedings of the 37th International Conference on Machine Learning},
  pages = 	 {3929--3938},
  year = 	 {2020},
  editor = 	 {III, Hal Daumé and Singh, Aarti},
  volume = 	 {119},
  series = 	 {Proceedings of Machine Learning Research},
  month = 	 {13--18 Jul},
  publisher =    {PMLR},
 
}

@article{zhang2023retrieve,
  title={Retrieve anything to augment large language models},
  author={Zhang, Peitian and Xiao, Shitao and Liu, Zheng and Dou, Zhicheng and Nie, Jian-Yun},
  journal={arXiv preprint arXiv:2310.07554},
  year={2023}
}

@article{gao2023retrieval,
  title={Retrieval-augmented generation for large language models: A survey},
  author={Gao, Yunfan and Xiong, Yun and Gao, Xinyu and Jia, Kangxiang and Pan, Jinliu and Bi, Yuxi and Dai, Yi and Sun, Jiawei and Wang, Haofen},
  journal={arXiv preprint arXiv:2312.10997},
  year={2023}
}

@inproceedings{caffagni2024wiki,
  title={Wiki-LLaVA: Hierarchical Retrieval-Augmented Generation for Multimodal LLMs},
  author={Caffagni, Davide and Cocchi, Federico and Moratelli, Nicholas and Sarto, Sara and Cornia, Marcella and Baraldi, Lorenzo and Cucchiara, Rita},
  booktitle={Proceedings of the IEEE/CVF Conference on Computer Vision and Pattern Recognition},
  pages={1818--1826},
  year={2024}
}

@inproceedings{xuretrieval,
  title={Retrieval meets Long Context Large Language Models},
  author={Xu, Peng and Ping, Wei and Wu, Xianchao and McAfee, Lawrence and Zhu, Chen and Liu, Zihan and Subramanian, Sandeep and Bakhturina, Evelina and Shoeybi, Mohammad and Catanzaro, Bryan},
  booktitle={The Twelfth International Conference on Learning Representations},
  year={2024}
}

@inproceedings{yoranmaking,
  title={Making Retrieval-Augmented Language Models Robust to Irrelevant Context},
  author={Yoran, Ori and Wolfson, Tomer and Ram, Ori and Berant, Jonathan},
  booktitle={The Twelfth International Conference on Learning Representations},
  year={2024}
}

@inproceedings{long2022retrieval,
  title={Retrieval augmented classification for long-tail visual recognition},
  author={Long, Alexander and Yin, Wei and Ajanthan, Thalaiyasingam and Nguyen, Vu and Purkait, Pulak and Garg, Ravi and Blair, Alan and Shen, Chunhua and van den Hengel, Anton},
  booktitle={Proceedings of the IEEE/CVF conference on computer vision and pattern recognition},
  pages={6959--6969},
  year={2022}
}

@inproceedings{hu2023reveal,
  title={Reveal: Retrieval-augmented visual-language pre-training with multi-source multimodal knowledge memory},
  author={Hu, Ziniu and Iscen, Ahmet and Sun, Chen and Wang, Zirui and Chang, Kai-Wei and Sun, Yizhou and Schmid, Cordelia and Ross, David A and Fathi, Alireza},
  booktitle={Proceedings of the IEEE/CVF conference on computer vision and pattern recognition},
  pages={23369--23379},
  year={2023}
}

@inproceedings{rao2023retrieval,
  title={Retrieval-based knowledge augmented vision language pre-training},
  author={Rao, Jiahua and Shan, Zifei and Liu, Longpo and Zhou, Yao and Yang, Yuedong},
  booktitle={Proceedings of the 31st ACM International Conference on Multimedia},
  pages={5399--5409},
  year={2023}
}

@article{blattmann2022retrieval,
  title={Retrieval-augmented diffusion models},
  author={Blattmann, Andreas and Rombach, Robin and Oktay, Kaan and M{\"u}ller, Jonas and Ommer, Bj{\"o}rn},
  journal={Advances in Neural Information Processing Systems},
  volume={35},
  pages={15309--15324},
  year={2022}
}

@inproceedings{sohl2015deep,
  title={Deep unsupervised learning using nonequilibrium thermodynamics},
  author={Sohl-Dickstein, Jascha and Weiss, Eric and Maheswaranathan, Niru and Ganguli, Surya},
  booktitle={International conference on machine learning},
  pages={2256--2265},
  year={2015},
  organization={PMLR}
}

@article{song2020denoising,
  title={Denoising diffusion implicit models},
  author={Song, Jiaming and Meng, Chenlin and Ermon, Stefano},
  journal={arXiv preprint arXiv:2010.02502},
  year={2020}
}

@article{ho2020denoising,
  title={Denoising diffusion probabilistic models},
  author={Ho, Jonathan and Jain, Ajay and Abbeel, Pieter},
  journal={Advances in neural information processing systems},
  volume={33},
  pages={6840--6851},
  year={2020}
}

@inproceedings{sheyninknn,
  title={kNN-Diffusion: Image Generation via Large-Scale Retrieval},
  author={Sheynin, Shelly and Ashual, Oron and Polyak, Adam and Singer, Uriel and Gafni, Oran and Nachmani, Eliya and Taigman, Yaniv},
  booktitle={The Eleventh International Conference on Learning Representations},
  year= {2023}
}

@inproceedings{zhang2023remodiffuse,
  title={Remodiffuse: Retrieval-augmented motion diffusion model},
  author={Zhang, Mingyuan and Guo, Xinying and Pan, Liang and Cai, Zhongang and Hong, Fangzhou and Li, Huirong and Yang, Lei and Liu, Ziwei},
  booktitle={Proceedings of the IEEE/CVF International Conference on Computer Vision},
  pages={364--373},
  year={2023}
}

@article{adolf2018rosettaantibodydesign,
  title={RosettaAntibodyDesign (RAbD): A general framework for computational antibody design},
  author={Adolf-Bryfogle, Jared and Kalyuzhniy, Oleks and Kubitz, Michael and Weitzner, Brian D and Hu, Xiaozhen and Adachi, Yumiko and Schief, William R and Dunbrack Jr, Roland L},
  journal={PLoS computational biology},
  volume={14},
  number={4},
  pages={e1006112},
  year={2018},
  publisher={Public Library of Science San Francisco, CA USA}
}

@article{dunbar2014sabdab,
  title={SAbDab: the structural antibody database},
  author={Dunbar, James and Krawczyk, Konrad and Leem, Jinwoo and Baker, Terry and Fuchs, Angelika and Georges, Guy and Shi, Jiye and Deane, Charlotte M},
  journal={Nucleic acids research},
  volume={42},
  number={D1},
  pages={D1140--D1146},
  year={2014},
  publisher={Oxford University Press}
}

@article{shanker2024unsupervised,
  title={Unsupervised evolution of protein and antibody complexes with a structure-informed language model},
  author={Shanker, Varun R and Bruun, Theodora UJ and Hie, Brian L and Kim, Peter S},
  journal={Science},
  volume={385},
  number={6704},
  pages={46--53},
  year={2024},
  publisher={American Association for the Advancement of Science}
}

@inproceedings{zheng2023structure,
  title={Structure-informed language models are protein designers},
  author={Zheng, Zaixiang and Deng, Yifan and Xue, Dongyu and Zhou, Yi and Ye, Fei and Gu, Quanquan},
  booktitle={International conference on machine learning},
  pages={42317--42338},
  year={2023},
  organization={PMLR}
}

@article{berman2000protein,
  title={The protein data bank},
  author={Berman, Helen M and Westbrook, John and Feng, Zukang and Gilliland, Gary and Bhat, Talapady N and Weissig, Helge and Shindyalov, Ilya N and Bourne, Philip E},
  journal={Nucleic acids research},
  volume={28},
  number={1},
  pages={235--242},
  year={2000},
  publisher={Oxford University Press}
}

@article{presta1992antibody,
  title={Antibody engineering},
  author={Presta, Leonard G},
  journal={Current Opinion in Structural Biology},
  volume={2},
  number={4},
  pages={593--596},
  year={1992},
  publisher={Elsevier}
}

@article{al1997standard,
  title={Standard conformations for the canonical structures of immunoglobulins},
  author={Al-Lazikani, Bissan and Lesk, Arthur M and Chothia, Cyrus},
  journal={Journal of molecular biology},
  volume={273},
  number={4},
  pages={927--948},
  year={1997},
  publisher={Elsevier}
}

@article{jumper2021highly,
  title={Highly accurate protein structure prediction with AlphaFold},
  author={Jumper, John and Evans, Richard and Pritzel, Alexander and Green, Tim and Figurnov, Michael and Ronneberger, Olaf and Tunyasuvunakool, Kathryn and Bates, Russ and {\v{Z}}{\'\i}dek, Augustin and Potapenko, Anna and others},
  journal={nature},
  volume={596},
  number={7873},
  pages={583--589},
  year={2021},
  publisher={Nature Publishing Group}
}

@inproceedings{rao2021msa,
  title={MSA transformer},
  author={Rao, Roshan M and Liu, Jason and Verkuil, Robert and Meier, Joshua and Canny, John and Abbeel, Pieter and Sercu, Tom and Rives, Alexander},
  booktitle={International Conference on Machine Learning},
  pages={8844--8856},
  year={2021},
  organization={PMLR}
}

@inproceedings{huanginteraction,
  title={Interaction-based Retrieval-augmented Diffusion Models for Protein-specific 3D Molecule Generation},
  author={Huang, Zhilin and Yang, Ling and Zhou, Xiangxin and Qin, Chujun and Yu, Yijie and Zheng, Xiawu and Zhou, Zikun and Zhang, Wentao and Wang, Yu and Yang, Wenming},
  booktitle={Forty-first International Conference on Machine Learning},
  year={2024}
}

@article{zhou2024antigen,
  title={Antigen-Specific Antibody Design via Direct Energy-based Preference Optimization},
  author={Zhou, Xiangxin and Xue, Dongyu and Chen, Ruizhe and Zheng, Zaixiang and Wang, Liang and Gu, Quanquan},
  journal={arXiv preprint arXiv:2403.16576},
  year={2024}
}

@article{zhou2015rapid,
  title={Rapid search for tertiary fragments reveals protein sequence--structure relationships},
  author={Zhou, Jianfu and Grigoryan, Gevorg},
  journal={Protein Science},
  volume={24},
  number={4},
  pages={508--524},
  year={2015},
  publisher={Wiley Online Library}
}

@article{shanehsazzadeh2023vitro,
  title={In vitro validated antibody design against multiple therapeutic antigens using generative inverse folding},
  author={Shanehsazzadeh, Amir and Alverio, Julian and Kasun, George and Levine, Simon and Khan, Jibran A and Chung, Chelsea and Diaz, Nicolas and Luton, Breanna K and Tarter, Ysis and McCloskey, Cailen and others},
  journal={bioRxiv},
  pages={2023--12},
  year={2023},
  publisher={Cold Spring Harbor Laboratory}
}

@article{shanehsazzadeh2023unlocking,
  title={Unlocking de novo antibody design with generative artificial intelligence},
  author={Shanehsazzadeh, Amir and Bachas, Sharrol and McPartlon, Matt and Kasun, George and Sutton, John M and Steiger, Andrea K and Shuai, Richard and Kohnert, Christa and Rakocevic, Goran and Gutierrez, Jahir M and others},
  journal={bioRxiv},
  pages={2023--01},
  year={2023},
  publisher={Cold Spring Harbor Laboratory}
}

@inproceedings{freyprotein,
  title={Protein Discovery with Discrete Walk-Jump Sampling},
  author={Frey, Nathan C and Berenberg, Dan and Zadorozhny, Karina and Kleinhenz, Joseph and Lafrance-Vanasse, Julien and Hotzel, Isidro and Wu, Yan and Ra, Stephen and Bonneau, Richard and Cho, Kyunghyun and others},
  booktitle={The Twelfth International Conference on Learning Representations},
  year={2023}
}

@article{gainza2020deciphering,
  title={Deciphering interaction fingerprints from protein molecular surfaces using geometric deep learning},
  author={Gainza, Pablo and Sverrisson, Freyr and Monti, Frederico and Rodola, Emanuele and Boscaini, Davide and Bronstein, Michael M and Correia, Bruno E},
  journal={Nature Methods},
  volume={17},
  number={2},
  pages={184--192},
  year={2020},
  publisher={Nature Publishing Group US New York}
}

@article{gainza2023novo,
  title={De novo design of protein interactions with learned surface fingerprints},
  author={Gainza, Pablo and Wehrle, Sarah and Van Hall-Beauvais, Alexandra and Marchand, Anthony and Scheck, Andreas and Harteveld, Zander and Buckley, Stephen and Ni, Dongchun and Tan, Shuguang and Sverrisson, Freyr and others},
  journal={Nature},
  volume={617},
  number={7959},
  pages={176--184},
  year={2023},
  publisher={Nature Publishing Group UK London}
}

@article{aguilar2022fragment,
  title={Fragment-based computational design of antibodies targeting structured epitopes},
  author={Aguilar Rangel, Mauricio and Bedwell, Alice and Costanzi, Elisa and Taylor, Ross J and Russo, Rosaria and Bernardes, Gon{\c{c}}alo JL and Ricagno, Stefano and Frydman, Judith and Vendruscolo, Michele and Sormanni, Pietro},
  journal={Science Advances},
  volume={8},
  number={45},
  pages={eabp9540},
  year={2022},
  publisher={American Association for the Advancement of Science}
}

@article{chothia1987canonical,
  title={Canonical structures for the hypervariable regions of immunoglobulins},
  author={Chothia, Cyrus and Lesk, Arthur M},
  journal={Journal of molecular biology},
  volume={196},
  number={4},
  pages={901--917},
  year={1987},
  publisher={Elsevier}
}

@article{dunbar2016anarci,
  title={ANARCI: antigen receptor numbering and receptor classification},
  author={Dunbar, James and Deane, Charlotte M},
  journal={Bioinformatics},
  volume={32},
  number={2},
  pages={298--300},
  year={2016},
  publisher={Oxford University Press}
}

@article{dauparas2022robust,
  title={Robust deep learning--based protein sequence design using ProteinMPNN},
  author={Dauparas, Justas and Anishchenko, Ivan and Bennett, Nathaniel and Bai, Hua and Ragotte, Robert J and Milles, Lukas F and Wicky, Basile IM and Courbet, Alexis and de Haas, Rob J and Bethel, Neville and others},
  journal={Science},
  volume={378},
  number={6615},
  pages={49--56},
  year={2022},
  publisher={American Association for the Advancement of Science}
}

@inproceedings{hsu2022learning,
  title={Learning inverse folding from millions of predicted structures},
  author={Hsu, Chloe and Verkuil, Robert and Liu, Jason and Lin, Zeming and Hie, Brian and Sercu, Tom and Lerer, Adam and Rives, Alexander},
  booktitle={International conference on machine learning},
  pages={8946--8970},
  year={2022},
  organization={PMLR}
}

@article{abanades2023immunebuilder,
  title={ImmuneBuilder: Deep-Learning models for predicting the structures of immune proteins},
  author={Abanades, Brennan and Wong, Wing Ki and Boyles, Fergus and Georges, Guy and Bujotzek, Alexander and Deane, Charlotte M},
  journal={Communications Biology},
  volume={6},
  number={1},
  pages={575},
  year={2023},
  publisher={Nature Publishing Group UK London}
}

@article{raybould2019five,
  title={Five computational developability guidelines for therapeutic antibody profiling},
  author={Raybould, Matthew IJ and Marks, Claire and Krawczyk, Konrad and Taddese, Bruck and Nowak, Jaroslaw and Lewis, Alan P and Bujotzek, Alexander and Shi, Jiye and Deane, Charlotte M},
  journal={Proceedings of the National Academy of Sciences},
  volume={116},
  number={10},
  pages={4025--4030},
  year={2019},
  publisher={National Acad Sciences}
}

@article{shirai1999h3,
  title={H3-rules: identification of CDR-H3 structures in antibodies},
  author={Shirai, Hiroki and Kidera, Akinori and Nakamura, Haruki},
  journal={FEBS letters},
  volume={455},
  number={1-2},
  pages={188--197},
  year={1999},
  publisher={Wiley Online Library}
}

@article{alford2017rosetta,
  title={The Rosetta all-atom energy function for macromolecular modeling and design},
  author={Alford, Rebecca F and Leaver-Fay, Andrew and Jeliazkov, Jeliazko R and O’Meara, Matthew J and DiMaio, Frank P and Park, Hahnbeom and Shapovalov, Maxim V and Renfrew, P Douglas and Mulligan, Vikram K and Kappel, Kalli and others},
  journal={Journal of chemical theory and computation},
  volume={13},
  number={6},
  pages={3031--3048},
  year={2017},
  publisher={ACS Publications}
}

@inproceedings{
villegas-morcillo2023guiding,
title={Guiding diffusion models for antibody sequence and structure co-design with developability properties},
author={Amelia Villegas-Morcillo and Jana Weber and Marcel Reinders},
booktitle={NeurIPS 2023 Generative AI and Biology (GenBio) Workshop},
year={2023},
}

@inproceedings{wuHTP,
 author = {Wu, Fang and Li, Stan Z.},
 booktitle = {Advances in Neural Information Processing Systems},
 editor = {A. Oh and T. Naumann and A. Globerson and K. Saenko and M. Hardt and S. Levine},
 pages = {31140--31157},
 publisher = {Curran Associates, Inc.},
 title = {A Hierarchical Training Paradigm for Antibody Structure-sequence Co-design},
 volume = {36},
 year = {2023}
}

@misc{kulytė2024improving,
      title={Improving Antibody Design with Force-Guided Sampling in Diffusion Models}, 
      author={Paulina Kulytė and Francisco Vargas and Simon Valentin Mathis and Yu Guang Wang and José Miguel Hernández-Lobato and Pietro Liò},
      year={2024},
      eprint={2406.05832},
      archivePrefix={arXiv},
      primaryClass={q-bio.QM},
}

@inproceedings{multHoo,
 author = {Hoogeboom, Emiel and Nielsen, Didrik and Jaini, Priyank and Forr\'{e}, Patrick and Welling, Max},
 booktitle = {Advances in Neural Information Processing Systems},
 editor = {M. Ranzato and A. Beygelzimer and Y. Dauphin and P.S. Liang and J. Wortman Vaughan},
 pages = {12454--12465},
 publisher = {Curran Associates, Inc.},
 title = {Argmax Flows and Multinomial Diffusion: Learning Categorical Distributions},
 volume = {34},
 year = {2021}
}

@inproceedings{
wang2023retrievalbased,
title={Retrieval-based Controllable Molecule Generation},
author={Zichao Wang and Weili Nie and Zhuoran Qiao and Chaowei Xiao and Richard Baraniuk and Anima Anandkumar},
booktitle={The Eleventh International Conference on Learning Representations },
year={2023},
}

@article{ho2019axial,
  title={Axial attention in multidimensional transformers},
  author={Ho, Jonathan and Kalchbrenner, Nal and Weissenborn, Dirk and Salimans, Tim},
  journal={arXiv preprint arXiv:1912.12180},
  year={2019}
}

@article{lin2023evolutionary,
  title={Evolutionary-scale prediction of atomic-level protein structure with a language model},
  author={Lin, Zeming and Akin, Halil and Rao, Roshan and Hie, Brian and Zhu, Zhongkai and Lu, Wenting and Smetanin, Nikita and Verkuil, Robert and Kabeli, Ori and Shmueli, Yaniv and others},
  journal={Science},
  volume={379},
  number={6637},
  pages={1123--1130},
  year={2023},
  publisher={American Association for the Advancement of Science}
}

@inproceedings{he2016deep,
  title={Deep residual learning for image recognition},
  author={He, Kaiming and Zhang, Xiangyu and Ren, Shaoqing and Sun, Jian},
  booktitle={Proceedings of the IEEE conference on computer vision and pattern recognition},
  pages={770--778},
  year={2016}
}
\bibliographystyle{iclr2025_conference}

\newpage
\appendix

\newcounter{Sfigure}
\setcounter{Sfigure}{1}
\renewcommand{\thefigure}{S\arabic{Sfigure}}

\section{Additional details}
\label{appA}

\subsection{Model details}
For feature dimensions, we set the single residue feature dimension to 128 and the pair feature dimension to 64. We leverage 6 IPA layers to capture geometry information. ESM2 650M is utilized in our model to create the embedding of antibody sequences, and the embedding dimension is 1280. 
In the local CDR-focus network, two layers of axial attention were used (two tied row self-attention and two column self-attention). The embedding dimension is 384, the hidden dimension is 1536, and number of attention heads is 6.

\subsection{Implementation details}
Our model was developed and executed within the PyTorch framework. For training, We chose the Adam optimizer with a learning rate of 0.0001, weight decay of 0.0, and momentum parameters beta1 and beta2 set to 0.9 and 0.999, respectively. To dynamically adjust the learning rate, we employed plateau as learning rate scheduler. When the validation loss plateaued, the learning rate was reduced by a factor of 0.8, with a minimum learning rate set to 5e-6. The scheduler's patience was set to 10 epochs. The batch size is 8 during training. We design 8 samples for each CDR in the test set. All experiments are run on a single RTX4090 GPU, with a memory storage of 24GB.

Due to the high variability and specificity of the CDRH3 region, and it is considered the most critical part in determining antigen-antibody binding. We conducted separate training for the sequence design of this region, adding and removing noise only for the CDRH3 region in each training iteration, with a total of 100,000 iterations. The other five regions, being more conserved, were trained together for a total of 250,000 iterations (approximately equivalent to 50,000 iterations per region). The reverse generation process time step t is set to 100.

\subsection{Implementation of structural retrieval}
\label{app_a.3}
The input consists of the backbone atom coordinates of each amino acid in the CDR region, forming a set of coordinate points $\mathcal{X}$. $\sT$ represents the structures of all proteins in the PDB database. $\sA$ represents a set of protein fragments representing CDR-like fragments corresponding to the input CDR structure. $\mathcal{C}$ represents the set of fragments of each structure with a length of m. $J$ represents a linear motif centered on residue $ j$ in the structure, with a length equal to the query fragment.

Assume that the coordinates of residue $j$ are aligned with the central residue of $\mathcal{X}$, and then compute the RMSD of $\mathcal{X}$ when aligned onto $\tau$. If the input contains discontinuous multiple structures, cRMSD will be the cumulative RMSD of these structures. MaxA, maxB, and maxC are three different upper-bound thresholds. These thresholds are selected to improve the speed and accuracy of the retrieval algorithm (For detailed proof, please refer to MASTER \citep{zhou2015rapid}).

\subsection{Implementation of CDR-like database constructing}
\label{appa.4}
To eliminate the computational overhead caused by structural retrieval during the model's training and inferencing, we followed previous work \citep{aguilar2022fragment} and initially executed the retrieval algorithm on all CDR structures of all antibodies to construct a CDR-like database.

Each of the CDR structures is used as a query to search for structurally similar motifs in the PDB database. The MASTER algorithm is used to match all CDRs against the entire PDB database to find CDR-like structures. This structural search is based on the Kabsch algorithm, using the RMSD of the $C\alpha$ coordinates. For CDR fragments of length 4, the RMSD threshold is 0.4, and the threshold is increased by 0.05 Å for each additional residue (with a maximum threshold set to 1.0 Å). 
In this way, we obtain a CDR-like fragments database corresponding to all CDR structures. Except for strictly filtering out results identical to real CDR sequences, no CDR sequence information was leaked in this process.

\subsection{Baseline details}
\label{appa.5}
\subsubsection{Traditional methods}
\textbf{Rosetta-Fixbb} \citep{adolf2018rosettaantibodydesign}  Rosetta-Fixbb can use energy functions for antibody CDR sequence design. Since DiffAb has already been proven to outperform it \citep{luo2022antigen,wuHTP} on sequence design task, we did not conduct additional comparisons.

\textbf{Grafting} To simulate the rational design commonly used in traditional antibody design methods, which often involves grafting CDR loops. For each CDR region, we directly selected the top-1 fragment(best) from the retrieval database for the structures in the test set and replaced the corresponding original CDR loop sequences with it.

\subsubsection{Deep learning methods}
\textbf{ProteinMPNN} \citep{dauparas2022robust} 
ProteinMPNN is a deep learning framework for protein sequence inverse folding. It leverages a message passing neural network to model the complex relationships between amino acids in a protein structure. We use the antibody's backbone structure as input and keep the sequences outside the CDR regions to be designed fixed. We design sequences for each CDR region separately. The sampling temperature is set to the default value of 0.1.

\textbf{Esm-IF1} \citep{hsu2022learning}
Esm-IF1 is a protein sequence inverse folding model trained on millions of AlphaFold2 predicted structures. We use the antibody's backbone structure as input and keep the sequences outside the CDR regions to be designed fixed. We design sequences for each CDR region separately. The sampling temperature is set to the value of 0.2.

\textbf{Diffab-fix} \citep{luo2022antigen}
Diffab is a diffusion model that can design sequences of CDR region with a fixed CDR backbone. It takes antigen-antibody framework context as condition to design CDR sequence. For a fair comparison, we retrained it with the default training configuration \textit{fixbb.yml}. 

\textbf{AbMPNN} \citep{dreyer2023inverse}
AbMPNN is fine-tuned by antibody structure data and predicted OAS (Observed Antibody Space ) structure data. Its model architecture is consistent with ProteinMPNN but achieves better performance in antibody inverse folding. We use the antibody's backbone structure as input and keep the sequences outside the CDR regions to be designed fixed. 
However, it is not open-sourced yet, so we evaluate it on its own test set. We design sequences for each CDR region separately. The sampling temperature is set to the default value of 0.1.
\setcounter{Sfigure}{1}
\subsection{Experiment on CDR-H3's length}
\begin{figure}[h]
    \centering
    \includegraphics[width=0.75\linewidth]{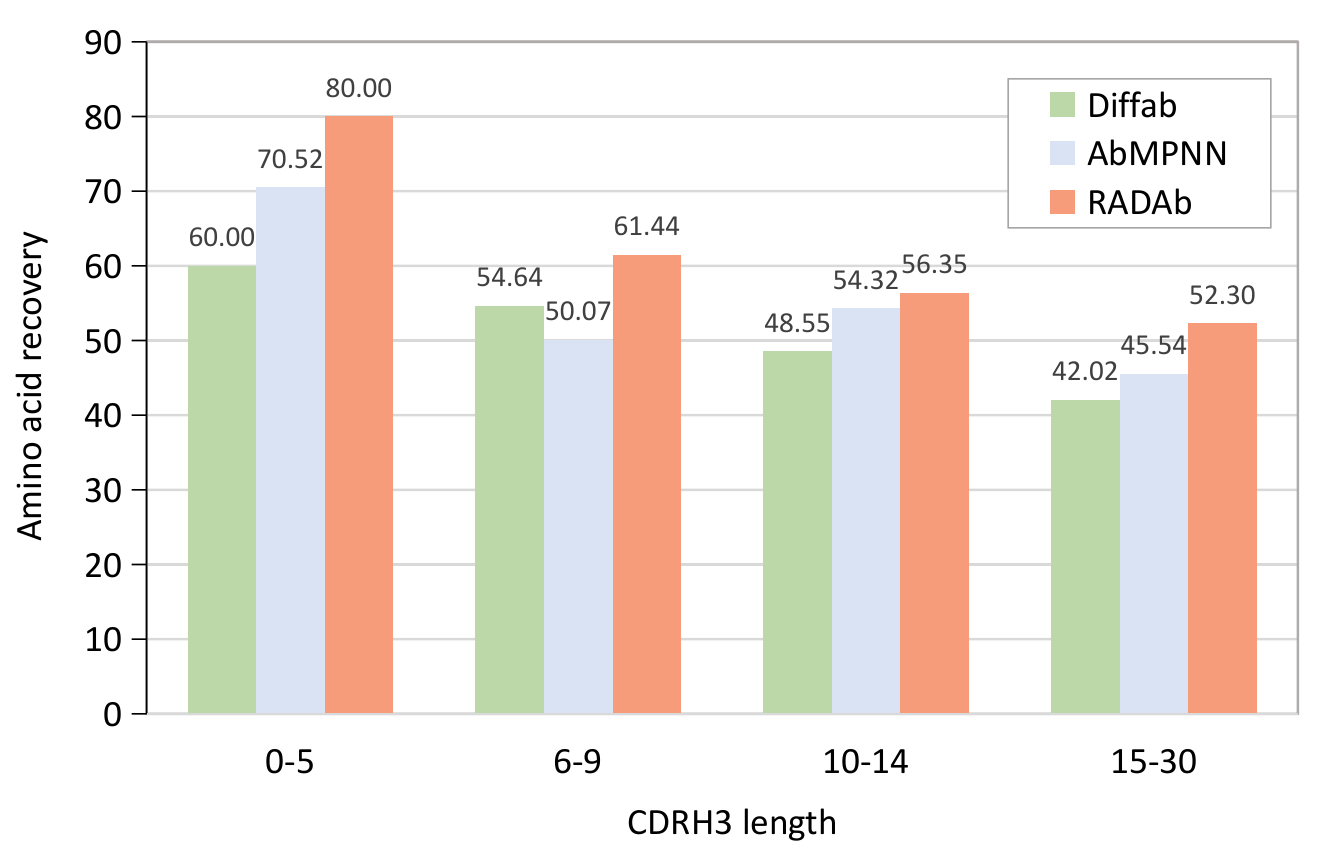}
    \caption{AAR distribution of different CDRH3 length}
    \label{fig_S1}
\end{figure}
We further evaluated each model's AAR across different CDR-H3 lengths. As shown in Figure \ref{fig_S1}, although the performance of all models decreases with increasing H3 length, our method still outperforms the others.

\subsection{Detailed results of antibody functionality optimization }
To further evaluate the model's performance in optimizing antibody functionality, we additionally assessed the $\Delta\Delta G_{seq}$ of the top-1 to top-3 structures generated by the model. The results are shown in Table \ref{energy_detail}, which further demonstrate that the antibodies optimized by our model exhibit improved functionality.

\begin{table}[]
    \centering
    \begin{tabular}{ccccccc}
            \toprule
             Method & {$\Delta$$\Delta$G$\downarrow$}  & $\Delta$$\Delta$G-seq $\downarrow$ & IMP-seq(\%) $\uparrow$ & F-top1 $\downarrow$ &F-top2 $\downarrow$&F-top3 $\downarrow$ \\
            \midrule
            Grafting &135.17 &40.22 &32.69&-&-&-\\
            \midrule
            ProteinMPNN & 127.14 & 24.72 & 35.51&-54.63&-45.58&-35.70 \\
            ESM-IF1 & 162.09 & 42.28 & 33.33&-62.65&-48.81&-33.86 \\
            Diffab-fix & 116.36 & 14.05 & 34.52&-62.51&-54.80&\textbf{-46.13} \\
            \midrule
            \textbf{RADAb} & \textbf{109.16} & \textbf{7.06} & \textbf{37.30}&\textbf{-69.30}&\textbf{-55.95}& -45.96  \\
            \bottomrule
        \end{tabular}
    \caption{Detailed results of antibody functionality optimization, F means functionality, which refers to $\Delta$$\Delta$G-seq.}
    \label{energy_detail}
\end{table}

\section{Structure reconstruction}
To reconstruct the antibody structure, we use ABodyBuilder2 \citep{abanades2023immunebuilder}, a deep learning model capable of predicting antibody light chain-heavy chain complexes. It is significantly faster than AlphaFold2 and offers higher prediction accuracy. We insert the designed CDR sequences into the antibody framework sequence, input it into ABodyBuilder2 to fold, and use OpenMM relax to obtain the structure corresponding to the new CDR sequences. Subsequently, we align the structure to the real antibody framework. Finally, we use the \textit{fastrelax} function in PyRosetta \citep{alford2017rosetta}, with the score function set to \textit{ref2015} and max iteration set to 1000, to relax the structure.

\section{Training and inference algorithm}
In this section, we provide a detailed algorithm for the training (Algorithm \ref{alg-train}) and inferencing (Algorithm  \ref{alg-sample}) processes.
\begin{algorithm}[h]
    \caption{Training Procedure of RADAb}
    \label{alg-train}
    \begin{algorithmic}[1]
        \State Coordinates set $\mathcal{X} = \{x_{k}\mid k\in  \left\{ 1,...,m \right\}\}$
        \State $\sA$ = $\text{Retrieval}(\mathcal{X})$
        
        \While{not convergence}
        \State $ t \sim \operatorname{Uniform}(1,..., T)$
        \State $ q\left(\text{s}_{j}^{t-1}\mid \text{s}_{j}^{t}, \text{s}_{j}^{0} \right) =\frac{ q\left(\mathrm{~s}_j^t \mid \mathrm{s}_j^{t-1}\right) \cdot q\left(\mathrm{~s}_j^{t-1} \mid \mathrm{s}_j^{0}\right)}{q\left(\mathrm{~s}_j^{t} \mid \mathrm{s}_j^{0}\right)}$    
        \State $S = S_{\text{fr}} \cup s_j^t $
        \State $e^t = E(S)$
        \State Context conditions  $\mathcal{C}\leftarrow\{\mathcal{R}^t, \mathcal{C}_{ab}, \mathcal{C}_{ag}\}$
        \State      $
                    p\left( s_{j}^{t-1} \mid \mathcal{C}, \sA \right) = \operatorname{Multinomial}\big[ F\left(\mathcal{C}, e^t \right) 
                    + G \left( F\left( \mathcal{C}, e^t \right), \sA \right) \big] \left[ j \right]$
        \State $    L_{\text{type\,\,}}^{t}=\mathbb{E}_{\mathcal{R}^t\sim p}\left[ \frac{1}{m}\sum_j{D}_{\text{KL}}\left( q\left( \text{s}_{j}^{t-1}\mid \text{s}_{j}^{t},\text{s}_{j}^{0} \right) ||p\left( \text{s}_{j}^{t-1}\mid \mathcal{C},\sA \right) \right) \right]$
        
            \State $ F\left(\mathcal{C}, e^t \right) 
                    , G\left( F\left( \mathcal{C}, e^t \right), \sA\right) \leftarrow \operatorname{Adam}\left(L_{\text{type\,\,}}^{t}\right)$
        \EndWhile
        \State \Return $F(\cdot), G(\cdot)$
    \end{algorithmic}
\end{algorithm}
\begin{algorithm}[h]
    \caption{Sampling Procedure of RADAb}
    \label{alg-sample}
    \begin{algorithmic}[1]
        \State $s_{j}^{T} \sim \operatorname{Uniform}(20)$
        \State Backbone coordinates set $\mathcal{X} = \{x_{k}\mid k\in  \left\{ 1,...,m \right\}\}$
        \State $\sA$ = $\text{Retrieval}(\mathcal{X})$
        \For{$t = T$ {\bfseries to} $1$}
            \State $S = S_\text{fr} \cup s_j^t $
            \State $e^t = E(S)$
            \State Context conditions  $\mathcal{C}\leftarrow\{\mathcal{R}^t, \mathcal{C}_{ab}, \mathcal{C}_{ag}\}$
            \State $p\left(s_j^{t-1}\right) =  \operatorname{Multinomial}\big[ F\left( \mathcal{C}, e^t\right) + G \left( F\left(  \mathcal{C}, e^t \right), \sA \right) \big]$
            \State sample $s_j^{t-1}$ from $p\left(s_j^{t-1}\right)$
            \State $R^{t-1} = \{s_j^{t-1} \mid j \in (a+1, \ldots, a+m)\}$
        \EndFor
        \State \Return $R^0$
    \end{algorithmic}
\end{algorithm}

\section{Case Study}
\begin{figure}[t]
    \centering
    \includegraphics[width=0.8\linewidth]{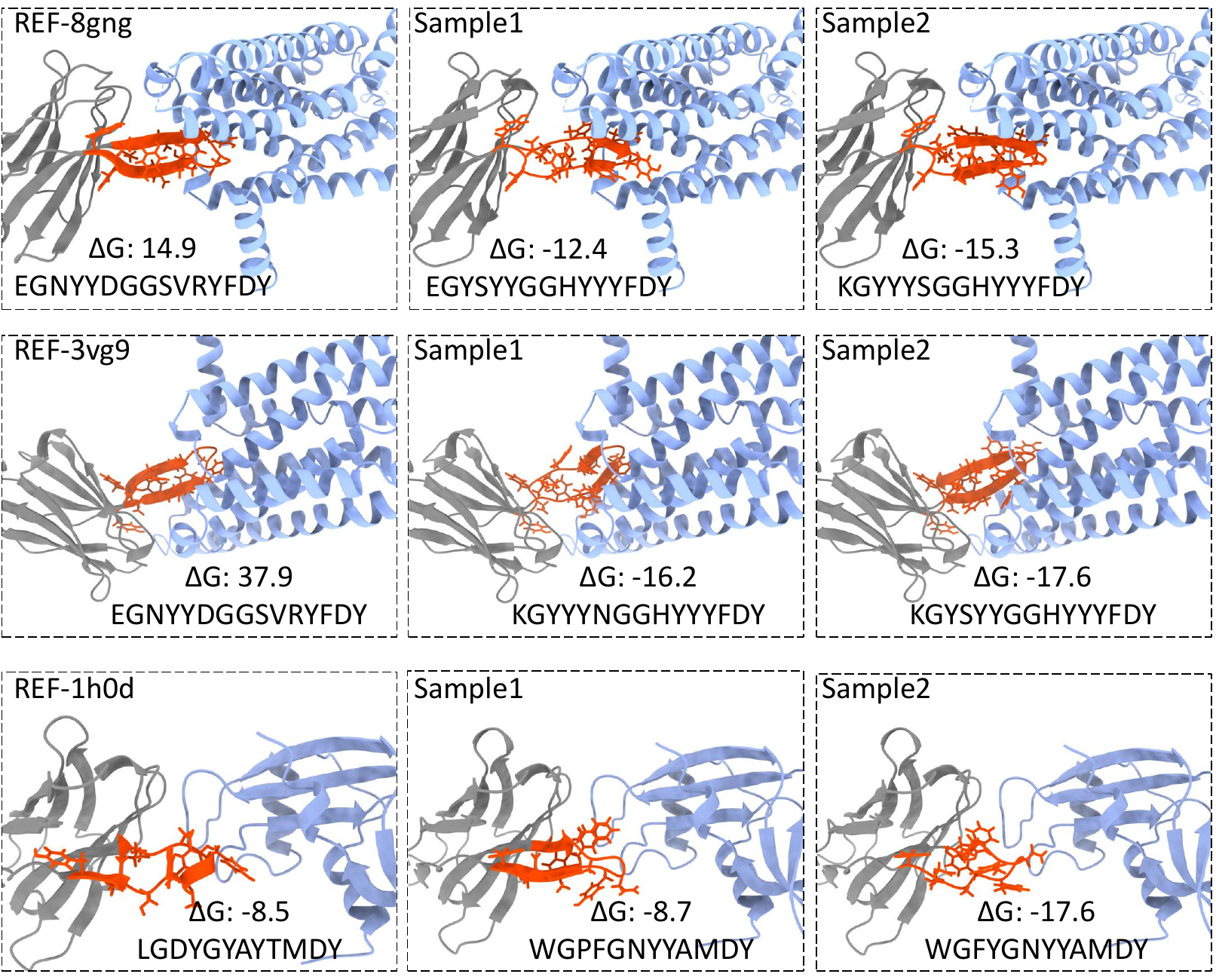}
    \caption{Optimized antibodies with lower binding energy. The gray parts represent the antibody framework, the red parts indicate the designed CDR regions, and the blue parts represent the antigen.}
    \label{case_app}
\end{figure}

We select a portion of the optimized antibodies in Figure \ref{case_app}. They achieved lower binding energy compared to the original antibody structures.
\setcounter{Sfigure}{2}

\section{The motivation for only considering sequence design}
\label{motivation}
Based on our observations, inverse folding represents a more practical scenario. Current structure-sequence co-design methods typically involve masking the CDR while retaining the presence of an antibody framework backbone, which represents a relatively uncommon use case. In most practical scenarios, we either have access to the full complex structure of the template antibody and the antigen, allowing us to perform inverse folding, or we lack a template molecule entirely, necessitating full atom and \textit{de novo}  design. This is also why researches from pharmaceutical companies and efforts involving in vitro experiments on antibody loop regions tend to focus more on designing antibodies through inverse folding, as evidenced by several recent studies \citep{shanehsazzadeh2023vitro,shanehsazzadeh2023unlocking,freyprotein,hoie2024antifold,shanker2024unsupervised}. This rationale underpins our decision to concentrate solely on this aspect in our work.

Another reason is that performing sequence-structure co-design while adhering to our retrieval-based approach would risk data leakage. This is because our retrieval process relies on the known structure of the CDR region, which is only possible when the CDR backbone is already known. At the same time, we recognize that epitope-specific full antibody \textit{de novo} design, including full-atom design, is highly valuable. We are actively developing retrieval systems for PPI (Protein-Protein Interaction) retrieval to avoid potential data leakage and enhance our model. 

\section{Limitations}
Due to the inherent limitations of the MASTER algorithms, the retrieved linear motifs may have structural issues. Despite careful screening and filtering, a tiny portion of the data might have lengths that differ from the original CDR loops or may even become discontinuous due to missing residues. These exceptional cases may have a negative impact on our model. We hope that advances in structural retrieval and improvements in alignment will jointly address this issue.

When constructing the CDR-like fragments database, searching the entire PDB database using all CDR structures from the SabDab database takes approximately 100 hours. Additionally, the ESM2 encoding used to capture antibody sequence evolution information and the axial attention focused on local CDR in the denoising network require more computational resources than typical diffusion methods.

It is important to note that applying current retrieval mechanism to structure and sequence co-design poses a significant risk of data leakage. Specifically, the retrieval process inherently relies on using the CDR structure information as a query, which practically transforms the task into one resembling inverse folding. However, one potential approach is to incorporate PPI (Protein-Protein Interaction) retrieval, and we are currently experimenting with MASIF\citep{gainza2020deciphering,gainza2023novo} for this purpose. This will be explored further as part of our future work. 

\end{document}